\documentclass{article}

\usepackage{PRIMEarxiv}

\usepackage[utf8]{inputenc} 
\usepackage[T1]{fontenc}    
\usepackage[colorlinks=true,linkcolor=blue,citecolor=blue,urlcolor=blue,filecolor=blue]{hyperref}       
\usepackage{url}            
\usepackage{booktabs}       
\usepackage{array}          
\usepackage{tabularx}       
\usepackage{amsfonts}       
\usepackage{nicefrac}       
\usepackage{microtype}      
\usepackage{lipsum}
\usepackage{fancyhdr}       
\usepackage{graphicx}       
\usepackage{subfigure}      
\usepackage{multirow}
\usepackage{amsmath,amssymb}
\usepackage{amsthm}
\usepackage{mathrsfs}
\usepackage{xcolor}
\usepackage{textcomp}
\usepackage{algorithm}
\usepackage{algorithmicx}
\usepackage{algpseudocode}
\usepackage{listings}
\usepackage{float}
\usepackage{subcaption}

\title{DCFFSNet: Deep Connectivity Feature Fusion Separation Network for Medical Image Segmentation
\thanks{\textit{\underline{Citation}}: 
\textbf{Zhang, M., Ye, X., Tang, R., Ding, H. DCFFSNet: Deep Connectivity Feature Fusion Separation Network for Medical Image Segmentation. }} 
}

\author{
  Mingda Zhang \\
  School of Software \\
  Yunnan University \\
  Kunming 650500, China\\
  \texttt{mingdazhang@ynu.edu.cn} \\
  \and
  Xun Ye \\
  School of Software \\
  Yunnan University \\
  Kunming 650500, China\\
  \texttt{xunye@ynu.edu.cn} \\
  \and
  Ruixiang Tang \\
  School of Software \\
  Yunnan University \\
  Kunming 650500, China\\
  \texttt{ruixiangtang@ynu.edu.cn} \\
  \and
  Haiyan Ding$^{*}$ \\
  School of Information \\
  Yunnan University \\
  Kunming 650500, China\\
  \texttt{dinghaiyan@ynu.edu.cn} \\
  \\
  \textit{$^{*}$Corresponding author}
}

\begin{document}
\maketitle

\begin{abstract}
Medical image segmentation leverages topological connectivity theory to enhance edge precision and regional consistency. However, existing deep networks integrating connectivity often forcibly inject it as an additional feature module, resulting in coupled feature spaces with no standardized mechanism to quantify different feature strengths. To address these issues, we propose DCFFSNet (Dual-Connectivity Feature Fusion-Separation Network). It introduces an innovative feature space decoupling strategy. This strategy quantifies the relative strength between connectivity features and other features. It then builds a deep connectivity feature fusion-separation architecture. This architecture dynamically balances multi-scale feature expression. Experiments were conducted on the ISIC2018, DSB2018, and MoNuSeg datasets. On ISIC2018, DCFFSNet outperformed the next best model (CMUNet) by 1.3\% (Dice) and 1.2\% (IoU). On DSB2018, it surpassed TransUNet by 0.7\% (Dice) and 0.9\% (IoU). On MoNuSeg, it exceeded CSCAUNet by 0.8\% (Dice) and 0.9\% (IoU). The results demonstrate that DCFFSNet exceeds existing mainstream methods across all metrics. It effectively resolves segmentation fragmentation and achieves smooth edge transitions. This significantly enhances clinical usability.
\end{abstract}


\section{Introduction}
\label{sec:introduction}

Medical image segmentation aims to perform pixel-level classification of organs or lesion regions in medical images, a task that poses significant challenges in terms of technical requirements and time costs for physicians. Traditional methods primarily rely on manually designed feature extraction. While they can effectively handle geometric deviations~\cite{ref1,ref2,ref3}, their performance in segmenting complex structures (such as organs or lesion regions with variable shapes and textures) is suboptimal.

In topology, connectivity describes how adjacent pixels are interrelated~\cite{ref4}. By defining the adjacency relationships between pixels, connectivity establishes spatial continuity constraints. This mathematical description effectively addresses two key shortcomings of traditional segmentation methods. First, by enhancing the modeling of correlations among edge pixels, it mitigates edge blurring caused by insufficient utilization of gradient information. Second, by establishing topological dependencies between regions, it improves the spatial consistency of segmentation results. Segmentation techniques based on connectivity models have achieved significant progress~\cite{ref5,ref6,ref7}. These connectivity-based techniques demonstrate stronger capabilities in handling images with complex structures, preserving internal structural continuity, reducing fragmentation in segmentation results, and optimizing segmentation edges~\cite{ref8}. Early applications of connectivity in deep learning primarily involved using traditional image segmentation methods for post-processing, such as applying the watershed algorithm~\cite{ref9} to refine the output of the DeepLab~\cite{ref10} model to enhance performance. These models did not truly integrate connectivity at a deeper level. However, with the advancement of connectivity technology, connectivity-based models have emerged. Traditional pixel-based segmentation mainly emphasizes categorical features, such as boundaries, whereas connectivity-based models exhibit superior capabilities in processing images with complex structures. In the field of deep learning for image segmentation, new applications leveraging connectivity continue to emerge.

Nevertheless, these connectivity-based networks often model connectivity as an additional feature injection and significantly enhance it within modules~\cite{ref8,ref11,ref12,ref13}. This approach may not be optimal because feature maps contain limited feature information. While extensive enhancement of connectivity features achieves good results in connectivity feature extraction, it simultaneously affects the acquisition of other features. These connectivity-based medical image segmentation networks lack a standardized method to measure the feature strength of different features in the feature space or to distinguish between them. These diverse features aggregate in the space, collectively forming the feature map.

Therefore, to address these issues:

\begin{enumerate}
\item This work decouples feature spaces by quantifying connectivity features relative to other features through standardized metrics.
\item We propose a Deep Connectivity Feature Fusion-Separation Network (DCFFSNet) based on feature space decoupling. By adaptively balancing the relative strength between connectivity scales and multi-scale features, it effectively resolves edge detail delineation challenges.
\end{enumerate}

\section{Related Work}
\label{sec:related_work}

\subsection{Connectivity Mask}

Traditional pixel-based segmentation methods primarily focus on categorical features, whereas connectivity-based models demonstrate superior performance in processing images with complex structures. After connectivity was widely adopted in classical image processing methods to describe topological properties~\cite{ref14,ref15}, it also found new applications in deep learning-based image segmentation~\cite{ref17,ref18,ref19,ref20}. Connectivity-based segmentation networks employ connectivity masks as labels. Typically, these masks consist of 8 channels (for 2D images, corresponding to the x and y axes) or 26 channels (for 3D images, corresponding to the x, y, and z axes)~\cite{ref21}, where each channel indicates whether a pixel in the original image belongs to the same class as its neighboring pixel in a specific direction.

A special case arises when converting standard masks to connectivity masks at image boundaries. For example, when performing upward connectivity operations on the top row of pixels, conventional methods treat non-existent neighboring positions as background (label 0). However, in this paper, such positions are assigned the same label as the pixel itself. This approach ensures that all eight channels in the connectivity mask's boundary regions retain their classification labels, thereby imposing stronger constraints on segmentation results near edges (where, under the original computation, at least three channels would be set to background).

\subsection{Bilateral Voting (BV) Module and Region-guided Channel Aggregation (RCA)}

After obtaining the connectivity mask, since the validation process requires converting the connectivity mask back to a general image mask, Yang et al.~\cite{ref16} were the first to propose a bilateral voting mechanism and a region-guided channel aggregation method to address the conversion between connectivity masks and general image masks.

The connectivity mask enhances consistency between adjacent pixels through the bilateral voting module. Because the connectivity mask only multiplies adjacent pixels of the same class---meaning the values of the two corresponding pixels in the connectivity mask should be closely related---the values in the corresponding two channels must belong to the same category to indicate consistent classification results. In bilateral voting, each pair of elements in the connectivity output $X^{C}$ is multiplied, and the result is assigned to these two elements. The resulting bilateral voting output $X^{B}$, referred to as the Bicon map, is treated as the connectivity mask $Y^{C}$ in the final output.

Each channel in the Bicon map displays the bilateral class probability distribution in a specific direction, reflecting the categorical relationship between a pixel and its neighboring pixels. The region-guided channel aggregation (RCA) method converts the Bicon map into a single-channel output probability through a mapping function.

\begin{equation}
X_{i}^{B}Mask(x,y) = \{f\{X_{9-i}^{B}(x,y)\}\}_{i=1}^{8}
\end{equation}

Here, $f$ can be any aggregation function. In this paper, we employ the $\max$ operation for channel aggregation, selecting the highest probability from the eight channels as the final predicted probability distribution.

\subsection{Attention Mechanism}

The Attention Mechanism~\cite{ref22} simulates the way humans focus on different regions with varying levels of attention and has been widely adopted in computer vision.

In image segmentation, the two most common and fundamental attention mechanisms are channel attention~\cite{ref23} and spatial attention~\cite{ref24}. In the channel attention mechanism, $w$ assigns different weights to each channel by capturing the feature strength and importance within the channels. In spatial attention, $w$ captures the feature strength and importance at different coordinates in the image or feature map, assigning distinct weights to each location.

\section{Method}
\label{sec:method}

\subsection{DCFFSNet Model}

\begin{figure}[H]
\centering
\includegraphics[width=\textwidth]{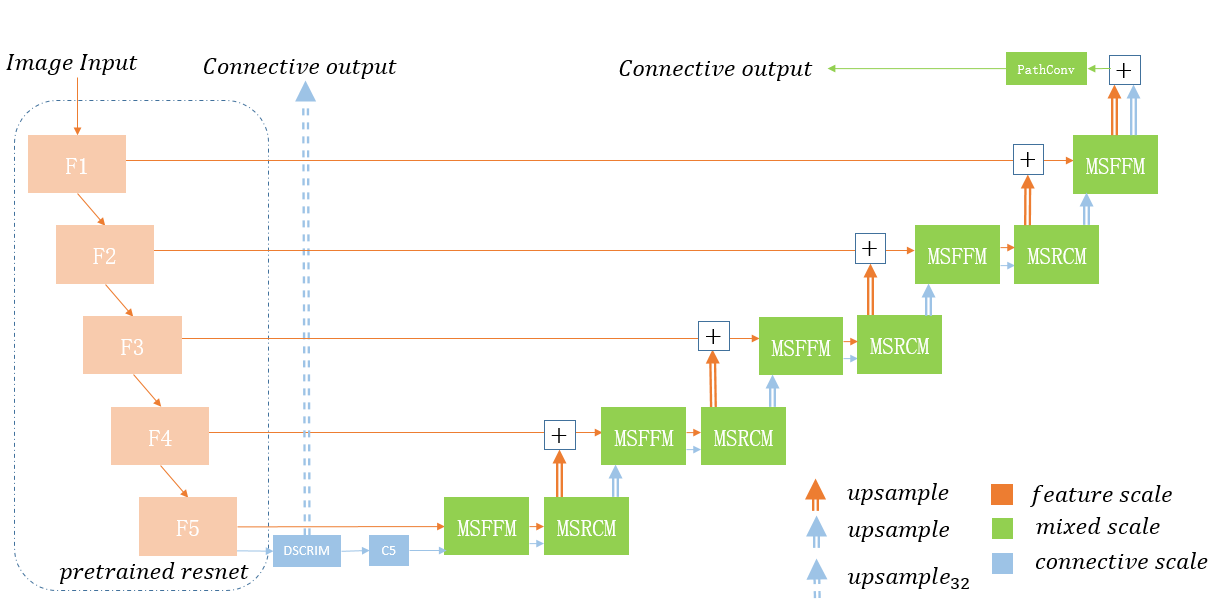}
\caption{Structure of the proposed DCFFSNet model, which consists of four components: the backbone network, the deeply supervised connectivity representation injection module, the multi-scale residual convolution module, and the directional convolution module.}
\label{fig:dcffsnet_structure}
\end{figure}

This paper proposes a DCFFSNet based on connectivity decoupling. The DCFFSNet adopts a typical U-shaped encoder-decoder architecture, and Figure~\ref{fig:dcffsnet_structure} illustrates the structure of DCFFSNet.

\subsection{Deeply Supervised Connectivity Representation Injection Module (DSCRIM)}

The DSCRIM is positioned at the bottleneck layer, where it injects connectivity features prominently into the feature space from the bottom layer through deep supervision to decouple connectivity features from classification features. This module significantly enhances the connectivity representation information of the feature map by employing rapid upsampling and connectivity grouping strategies, thereby capturing the connectivity scale features at the bottom layer.

\begin{figure}[H]
\centering
\includegraphics[width=\textwidth]{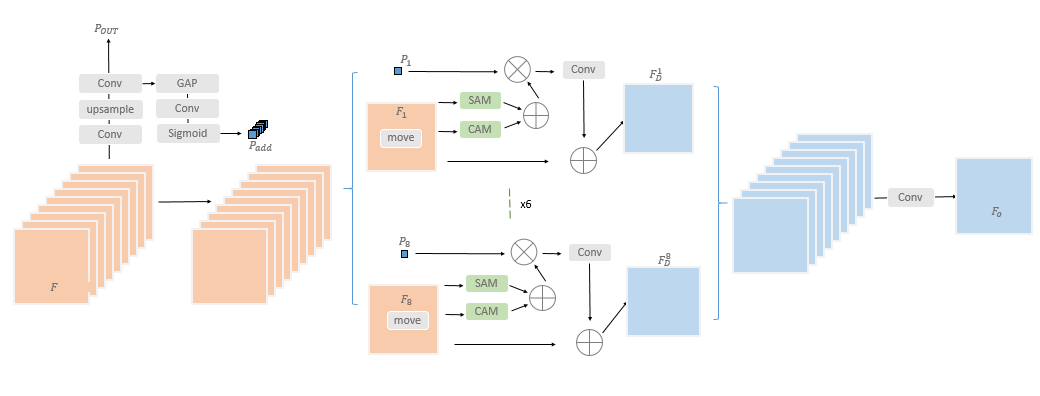}
\caption{Structure of DSCRIM}
\label{fig:dscrim}
\end{figure}

Figure~\ref{fig:dscrim} illustrates the detailed structure of the DSCRIM. Specifically, the DSCRIM first performs multi-fold upsampling on the bottom-layer feature map to generate the corresponding deep supervision output and injects the connectivity representation prominently into the network. Subsequently, it learns connectivity features at the channel level through the connectivity grouping strategy, further optimizing feature representation.

First, the module takes the input bottom-layer feature map $F_5$, performs convolution and multi-fold upsampling (Upsample\_32), and then applies a $1 \times 1$ convolution to obtain the output $P_{OUT}$ of the original size. This output is used to inject prominent connectivity representations into the bottom layer through deep supervision, effectively capturing global connectivity information in the feature map. Next, global average pooling (GAP) is applied to $P_{OUT}$, and the number of channels is adjusted to match the original input via a $1 \times 1$ convolution. After passing through an activation function, a set of connectivity features $P_{add}$ is generated to represent the connectivity relationships between channels.

\begin{equation}
P_{out} = Conv_{1 \times 1}(Upsample_{32}(Conv_{1 \times 1}(F_5)))
\end{equation}

$P_{add}$ is divided into 8 groups, each corresponding to connectivity features $P_i$ in different directions. The input feature map $F_5$ is also divided into 8 groups, and a shift operation is applied to each group to capture spatial dependencies in different directions, resulting in $X_i$.

Next, the module performs connectivity representation fusion for each group of features $X_i$ and $C_i$ through the following steps:

\begin{enumerate}
\item Pass $X_i$ through the spatial attention module (SAM) and the channel attention module (CAM) to capture long-range spatial dependencies and inter-channel interaction information, respectively.
\item Add the outputs of spatial attention and channel attention to obtain a fused feature representation combining both spatial and channel attention.
\item Multiply the fused features with the connectivity representation $P_i$ element-wise along the channel dimension to selectively enhance or suppress features with specific directional information.
\item Further optimize the processed features using a $1 \times 1$ convolution and add them to the original features to retain the original information, resulting in the intermediate feature $F_D$.
\end{enumerate}

\begin{equation}
F_D^i = X_i + Conv_{1 \times 1}(SpatialAtt(X_i) + ChannelAtt(X_i) \odot P_i)
\end{equation}

Here, $\odot$ denotes element-wise multiplication, $SpatialAtt$ and $ChannelAtt$ represent spatial attention and channel attention, respectively.

Finally, the 8 groups of processed features are concatenated to restore the original dimensions, and a $1 \times 1$ convolution is applied to decode the feature information, yielding the final bottom-layer connectivity feature output $C_5$.

\begin{equation}
C_5 = Conv_{1 \times 1}(cat(F_D^1, \ldots, F_D^8))
\end{equation}

\subsection{Multi-Scale Feature Fusion Module (MSFFM)}

The MSFFM fuses connectivity scales and feature scales of different dimensions to obtain the next dimension's connectivity scale features or hybrid features. It decouples the connectivity space and feature space through a self-attention mechanism.

\begin{figure}[H]
\centering
\includegraphics[width=0.8\textwidth]{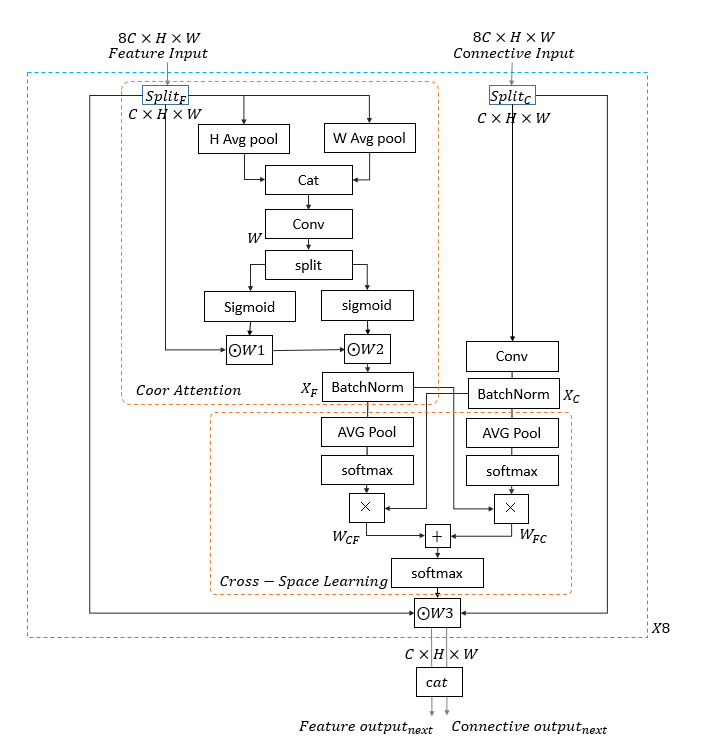}
\caption{Structure of MSFFM}
\label{fig:msffm}
\end{figure}

Figure~\ref{fig:msffm} illustrates the structure of the MSFFM. The module significantly enhances the spatial location information and global contextual dependencies of the connectivity scale feature map by fusing two types of inputs: feature scales and connectivity scales.

For feature-scale inputs, they are divided into eight groups, with the features in each group referred to as group features. The features undergo average pooling along both the horizontal and vertical directions. This process captures long-range dependencies in the horizontal dimension while retaining positional information in the vertical dimension. Subsequently, the pooled results from both directions are concatenated and encoded into intermediate weights using a convolution operation $W$.

\begin{equation}
W = Conv(cat(Avg_H(F_i), Avg_W(F_i)))
\end{equation}

Subsequently, crop $W$ along the spatial direction, and reshaped back to the original structure (the horizontal and vertical directions). Finally, a gating mechanism is used to generate the weight representation. After multiplying with the input $F_i$, the intermediate features at the feature scale are obtained through normalization.

\begin{equation}
X_F = BatchNormalize(F_i \odot Sigmoid(split(W)))
\end{equation}

For connectivity scale input $C$, divide it into 8 groups, each group is characterized by $C_i$. The difference is that here $3 \times 3$ $Conv$ is used to get the intermediate feature of connectivity scale $X_C$:

\begin{equation}
X_C = BatchNormalize(Conv_{3 \times 3}(C_i))
\end{equation}

After obtaining the intermediate features between the connectivity scale and the feature scale $X_C, X_F$, learn about both across space. The specific operation is as follows:

On features $X_F$, the 2D global average pooling layer is used to encode the global spatial information, and then $Softmax$ get the normalized channel descriptor $W_F$. Multiply $W_F$ with features $X_C$. That is, the weighted sum of all channel features at each location is obtained to obtain the global spatial attention representation in the connectivity scale $W_{FC}$.

\begin{equation}
W_F, W_{FC} = reshape(Softmax(Avg(X_F))), reshape(W_F \bullet reshape(X_C))
\end{equation}

Similar to the above operation, processing $X_C$ obtain global spatial attention representation on the feature scale $W_{CF}$.

Finally, the two kinds of spatial attention are aggregated and the final weight representation $W_a$ is obtained using the gating mechanism. The two inputs $F$ and $C$ are calibrated to get the outputs $C_{next}$ and $F_{next}$.

\begin{equation}
C_{next}, F_{next} = sigmoid(W_{CF} + W_{FC}) \bullet (C, F)
\end{equation}

Since the above operations are performed in all groups, the resulting output feature $C_{next}$ still has the original grouping structure of the connectivity scale and can be invoked repeatedly in multiple layers.

In particular, in the last layer of the model, the model uses the final output of a mixture of both.

\subsection{Multi-Scale Residual Convolution Module (MSRCM)}

MSRCM performs multi-scale feature extraction in up-sampling.

\begin{figure}[H]
\centering
\includegraphics[width=0.8\textwidth]{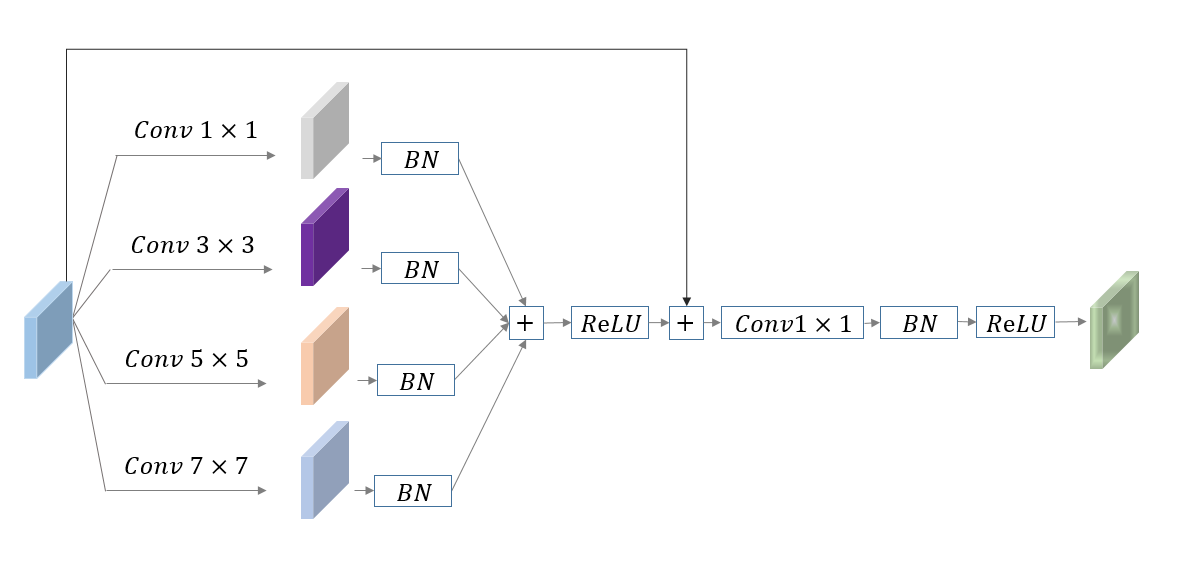}
\caption{Structure of MSRCM}
\label{fig:msrcm}
\end{figure}

Figure~\ref{fig:msrcm} shows the structure of MSRCM. MSRCM extracts multi-scale features through convolution kernels of different sizes and alleviates the problem of gradient disappearance through residual connection, thus improving the performance of the model.

\begin{equation}
Y = ReLU(BN(Conv_{1 \times 1}(X + ReLU(\sum_{i=1,3,5,7} BN(Conv_{i \times i}(X))))))
\end{equation}

MSRCM uses four convolutional kernels of different sizes to extract input features X to obtain multi-scale feature information. Then, these features are normalized by BN and added together for fusion. After ReLU operation, they are residual, and then re-encoded with a $1 \times 1$ convolutional kernel to obtain output Y.

\subsection{Directional Convolution (PConv)}

\begin{figure}[H]
\centering
\includegraphics[width=0.8\textwidth]{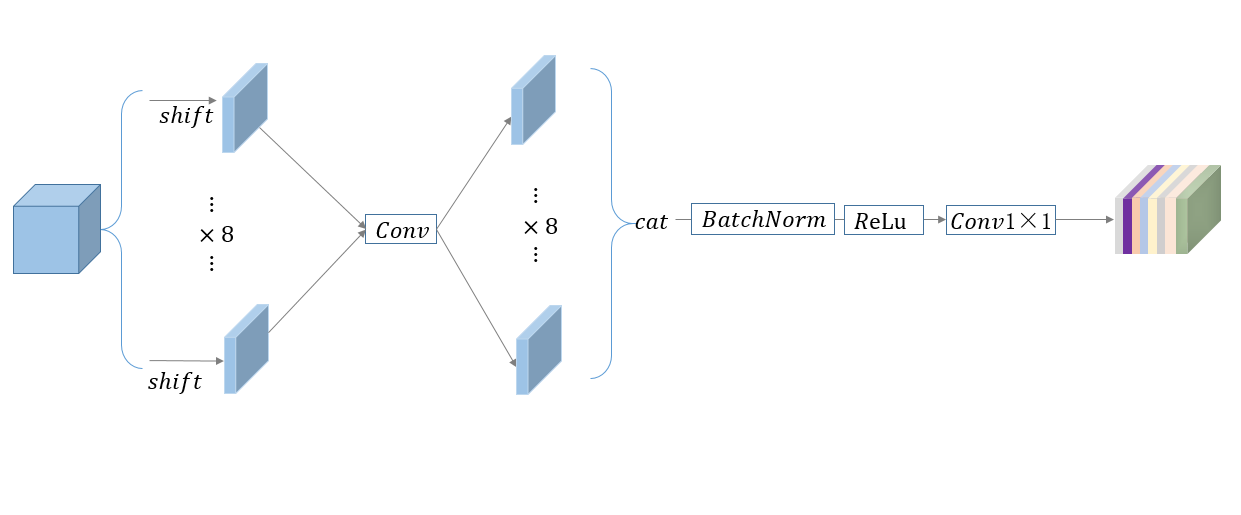}
\caption{Structure of Directional Convolution (PConv)}
\label{fig:pconv}
\end{figure}

The directional convolution method PConv (Path Convolution) optimizes the segmentation effect of the connectivity mask by grouping and shifting, transforming the output from the highest layer into the final prediction. Figure~\ref{fig:pconv} illustrates the structure of PConv. PConv groups and shifts the feature maps, applies a completely identical convolution kernel within each group, concatenates the results, and encodes them to achieve the interpretability of the connectivity mask.

\begin{equation}
Y = Conv_{1 \times 1}(ReLU(BN(cat(Conv_{3 \times 3}(shift(X_i))))))
\end{equation}

PConv divides the input feature map X into eight groups, performs shift operations corresponding to the connectivity direction within each group, and applies a $3 \times 3$ convolution kernel for convolution. The results are then reassembled to their original size and decoded using a $1 \times 1$ convolution kernel to produce the final model prediction output.

\subsection{Loss Function}

This paper considers the connectivity output $X^C$ as the core output of the model. In Section 2.1, it is described that the connectivity mask can be derived from the regular mask X through a simple multiplication operation. In Section 2.2, the connectivity output $X^C$ is obtained through bilateral voting to generate the dual connectivity mask $X^B$, which is then transformed into the standard mask $X$ via RCA.

Therefore, considering the above different masks, the loss function of the model is defined as follows:

\begin{align}
L &= L_{MainBCE} + 0.2 * L_{BBCE} + 0.8 * L_{CBCE} \\
L_{MainBCE} &= L_{BCE}(X, Y) = X * Y + (1 - X) * (1 - Y) \\
L_{BBCE} &= L_{BCE}(X^B, Y^C) = X^B * Y^C + (1 - X^B) * (1 - Y^C) \\
L_{CBCE} &= L_{BCE}(X^C, Y^C) = X^C * Y^C + (1 - X^C) * (1 - Y^C)
\end{align}

Among them, the superscript B refers to the dual pass mask, C refers to the connectivity mask, and the superscript without refers to the standard mask.

\begin{equation}
Loss = 0.2 * L(output1) + L(output2)
\end{equation}

The model proposed in this paper has two outputs, where $output1$ is the intermediate output of deep supervised connectivity representation injection, and $output2$ is the final output of the model.

\section{Experiments}
\label{sec:experiments}

\subsection{Dataset}

In order to verify the performance of the model, we used three different datasets for experiments, including ISIC2018 dataset~\cite{ref25}, DSB2018 dataset~\cite{ref26} and MoNuSeg dataset~\cite{ref27}.

\subsection{Comparative Experiments}

In order to verify the effectiveness of the proposed DCFFSNet network, we selected more classical medical image segmentation networks U-Net~\cite{ref28}, UNet++~\cite{ref29}, AttUNet~\cite{ref30}, TransUNet~\cite{ref31}, CMUNet~\cite{ref32}, CSCAUNet~\cite{ref33} for comparison. In this paper, the Dice Similarity Coefficient (DSC) and Intersection over Union (IoU) were selected as the core quantitative evaluation metrics.

Table~\ref{tab:comparison} shows the model performance of DCFFSNet under different datasets, where $\pm$ represents the standard deviation in a five-fold cross-validation environment, which reflects the stability of the model to some extent. The experimental results show that our model is the best.

\begin{table}[H]
\centering
\caption{Comparative experimental results of DCFFSNet}
\label{tab:comparison}
\resizebox{\textwidth}{!}{%
\begin{tabular}{lccccccccc}
\toprule
\multirow{2}{*}{Model} & \multirow{2}{*}{Year} & \multicolumn{2}{c}{ISIC2018} & \multicolumn{2}{c}{DSB2018} & \multicolumn{2}{c}{MoNuSeg} & \multirow{2}{*}{FLOPs} & \multirow{2}{*}{Params} \\
\cmidrule(lr){3-4} \cmidrule(lr){5-6} \cmidrule(lr){7-8}
& & IoU(\%) & Dice(\%) & IoU(\%) & Dice(\%) & IoU(\%) & Dice(\%) & & \\
\midrule
UNet~\cite{ref28} & 2015 & 79.8 $\pm$ 0.7 & 86.9 $\pm$ 0.8 & 83.8 $\pm$ 0.3 & 90.5 $\pm$ 0.2 & 63.1 $\pm$ 0.8 & 76.6 $\pm$ 0.7 & 50.166G & \textbf{34.527M} \\
UNet++~\cite{ref29} & 2018 & 79.9 $\pm$ 0.1 & 87.0 $\pm$ 0.2 & 84.5 $\pm$ 0.1 & 91.0 $\pm$ 0.1 & 63.7 $\pm$ 0.6 & 76.9 $\pm$ 0.5 & 106.162G & 36.630M \\
AttUNet~\cite{ref30} & 2019 & 81.8 $\pm$ 0.1 & 88.7 $\pm$ 0.2 & 84.1 $\pm$ 0.1 & 90.7 $\pm$ 0.1 & 64.1 $\pm$ 0.8 & 77.3 $\pm$ 0.8 & 51.015G & 34.879M \\
TransUNet~\cite{ref31} & 2021 & 80.9 $\pm$ 0.8 & 86.9 $\pm$ 0.2 & 84.7 $\pm$ 0.2 & 91.2 $\pm$ 0.3 & 65.7 $\pm$ 0.7 & 78.2 $\pm$ 0.7 & 32.238G & 93.231M \\
CMUNet~\cite{ref32} & 2023 & 82.2 $\pm$ 0.3 & 88.8 $\pm$ 0.3 & 83.9 $\pm$ 0.2 & 90.5 $\pm$ 0.2 & 66.1 $\pm$ 0.7 & 78.5 $\pm$ 0.8 & 69.866G & 49.932M \\
CSCAUNet~\cite{ref33} & 2024 & 82.0 $\pm$ 0.4 & 88.5 $\pm$ 0.4 & 84.4 $\pm$ 0.3 & 90.9 $\pm$ 0.3 & 66.4 $\pm$ 0.5 & 78.8 $\pm$ 0.6 & \textbf{10.517G} & 35.275M \\
DCFFSNet (Ours) & - & \textbf{83.5 $\pm$ 0.3} & \textbf{90.0 $\pm$ 0.2} & \textbf{85.4 $\pm$ 0.1} & \textbf{92.1 $\pm$ 0.1} & \textbf{67.2 $\pm$ 0.9} & \textbf{79.7 $\pm$ 0.9} & 21.732G & 52.717M \\
\bottomrule
\end{tabular}
}
\end{table}

By analyzing and comparing the experimental metrics, the DCFFSNet network proposed in this paper achieved the best Dice coefficient and IoU scores across three datasets. In the ISIC2018 dataset, the model scored 83.5 $\pm$ 0.3 in IoU and 90.0 $\pm$ 0.2 in Dice coefficient, which are 1.3\% and 1.2\% higher than the second-best CMUNet, respectively. In the DSB2018 dataset, the model scored 85.4\% in IoU and 92.1\% in Dice coefficient, which are 0.9\% and 0.7\% higher than the second-best TransUNet. In the MoNuSeg dataset, the model scored 67.2\% in IoU and 79.7\% in Dice coefficient, which are 0.8\% and 0.9\% higher than the second-best CSCAUNet.

DCFFSNet achieves an effective balance between computational efficiency and model performance. With a computational load of only 21.732 GFLOPs, it performs significantly lower than models like U-Net, UNet++, and CMUNet, while marginally exceeding CSCAUNet. This indicates reduced computational resource requirements during inference, enabling more efficient segmentation tasks---particularly suitable for deployment in computation-limited scenarios. Regarding parameters (52.717M), although DCFFSNet doesn't attain the lowest count, it delivers high segmentation performance without substantially increasing parametric burden. This demonstrates its optimal balance between model complexity and effectiveness. Collectively, while DCFFSNet doesn't dominate in absolute computational/parametric metrics, its lower computational load enhances deployment feasibility. The architecture demonstrates promise as an efficient segmentation model by maintaining favorable performance-efficiency trade-offs.

\begin{figure}[H]
\centering
\includegraphics[width=\textwidth]{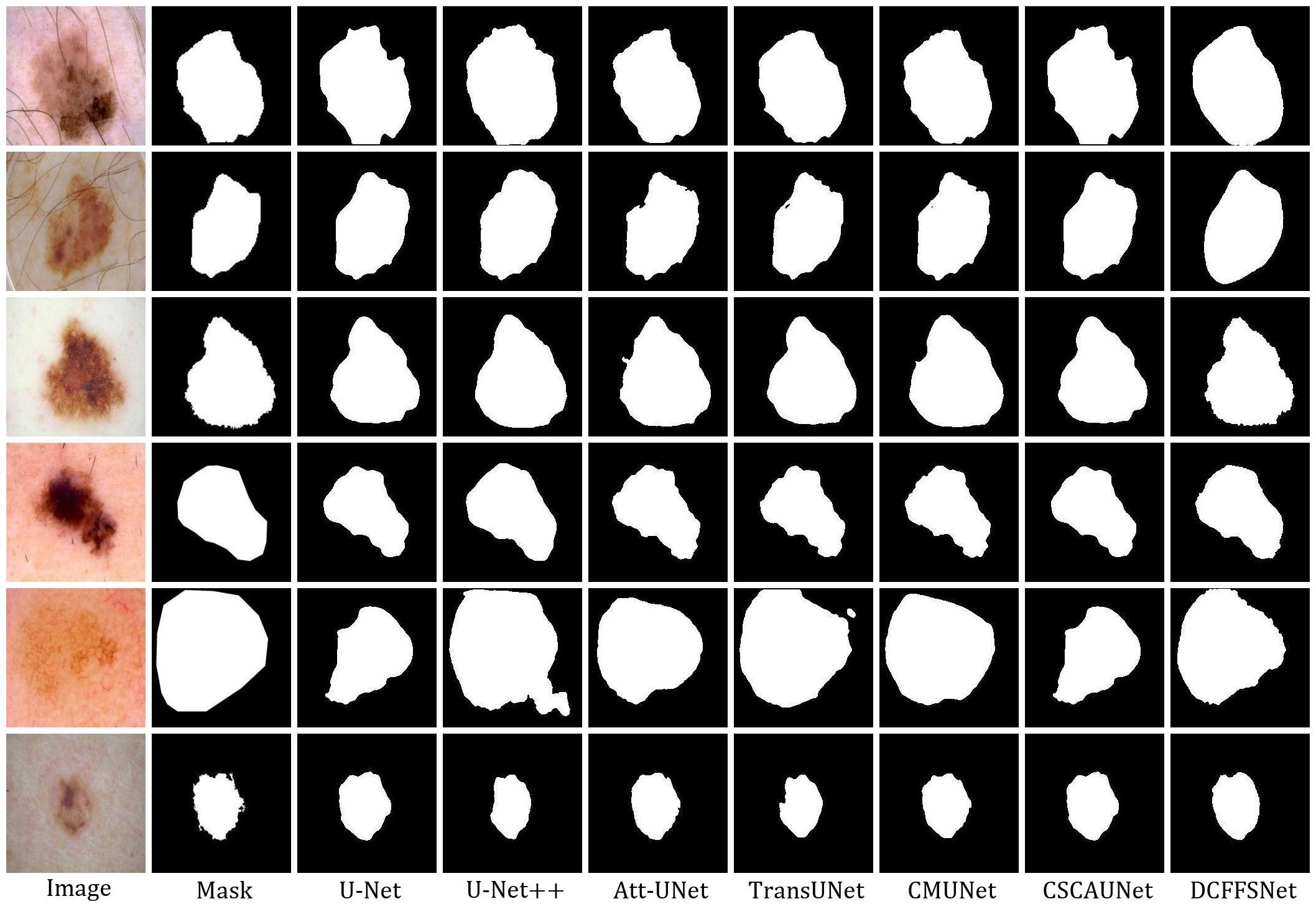}
\caption{Visualization reference of image comparison experiment of ISIC2018 dataset}
\label{fig:isic_results}
\end{figure}

\begin{figure}[H]
\centering
\includegraphics[width=\textwidth]{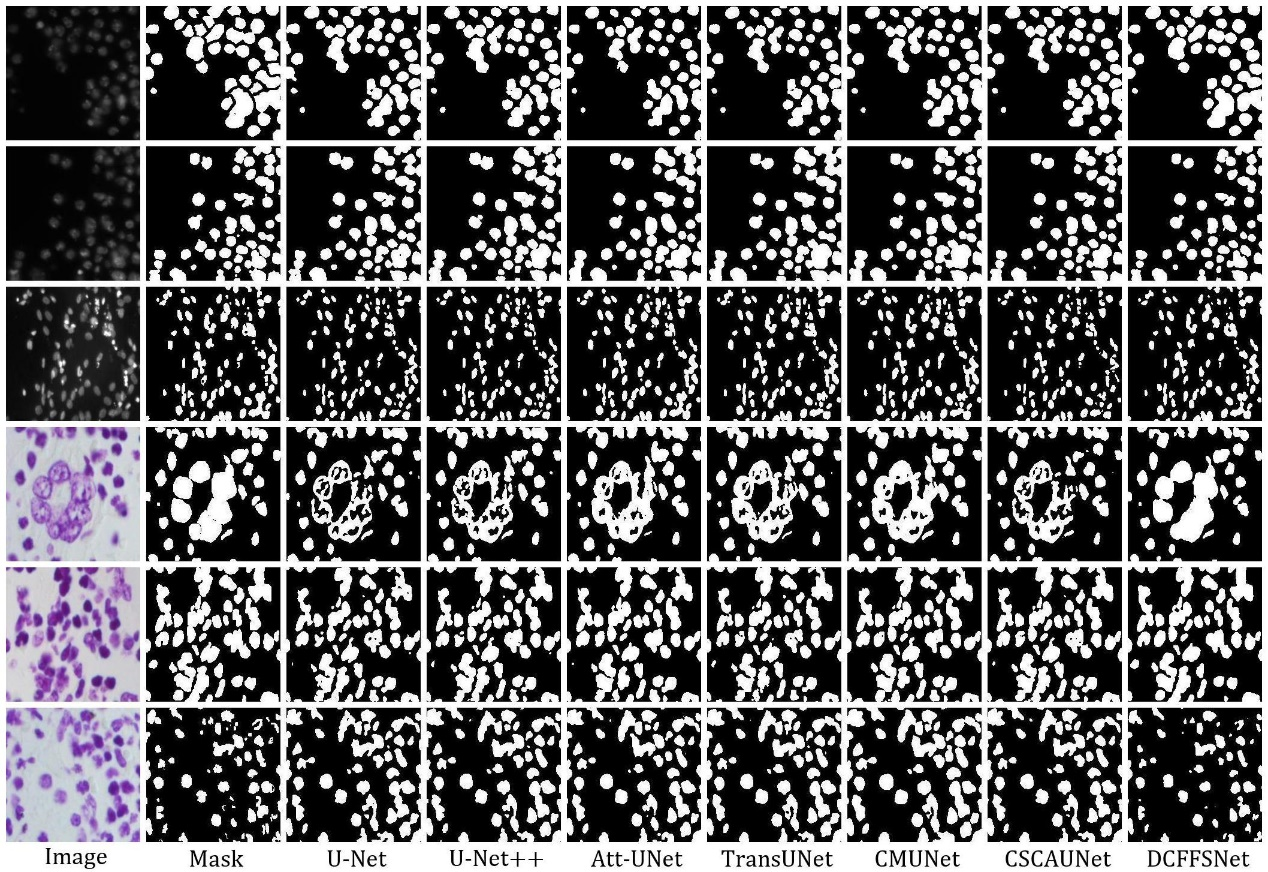}
\caption{Visualization reference of DSB2018 dataset comparison}
\label{fig:dsb_results}
\end{figure}

\begin{figure}[H]
\centering
\includegraphics[width=\textwidth]{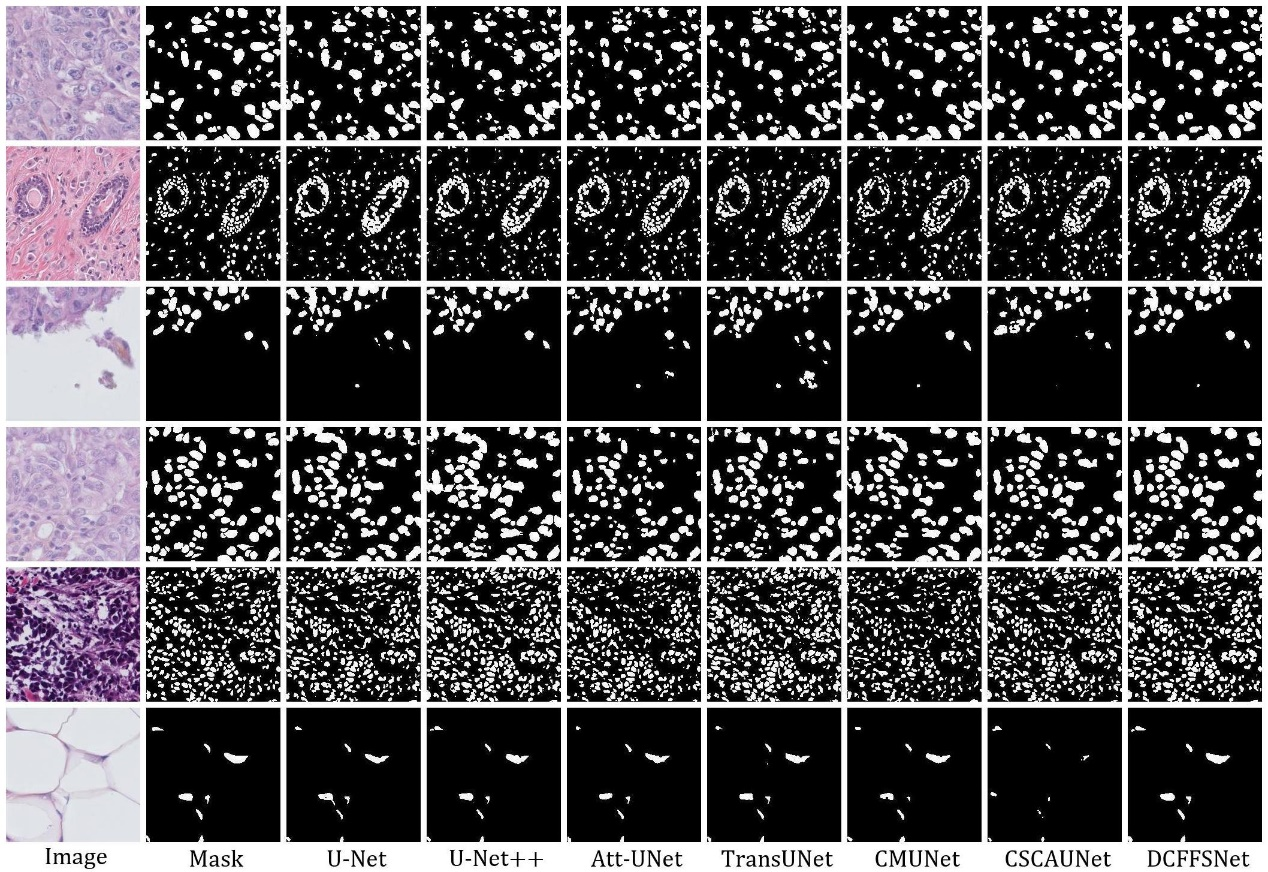}
\caption{Visual reference of image comparison experiment on MoNuSeg dataset}
\label{fig:monuseg_results}
\end{figure}

The visualization results in Figures~\ref{fig:isic_results},~\ref{fig:dsb_results}, and~\ref{fig:monuseg_results} demonstrate that DCFFSNet significantly outperforms comparative models including UNet and UNet++ in edge refinement, internal topology preservation, and overall segmentation quality. Unlike other models that frequently suffer from edge blurring, boundary misalignment, and internal discontinuities, DCFFSNet achieves precise edge localization with sharp contours highly consistent with ground-truth annotations while effectively maintaining intra-region connectivity and eliminating discontinuity artifacts, resulting in optimal morphological coherence and detail representation. By combining superior segmentation accuracy with operational efficiency through low computational load and moderate parameters, DCFFSNet establishes the best performance-efficiency balance among all evaluated models, highlighting its strong potential as an efficient segmentation solution and demonstrating significant practical value for flexible deployment in real-world medical image analysis applications.

\vspace{-15pt}

\vspace{-15pt}

\subsection{Ablation Experiments}

In order to verify the effectiveness of the proposed method and strictly control the variables, this section still selects ISIC2018 dataset, DSB2018 dataset and MoNuSeg dataset for ablation experiments.

Table~\ref{tab:ablation} shows the results of the ablation experiment. In this paper, the ablation experiment is carried out by removing or replacing modules. The specific network structure design is as follows:

\begin{itemize}
\item DCFFSNet: Deep connectivity feature fusion and separation network proposed in this chapter.
\item w/o DS: The deep supervision connectivity representation injection module DSCRIM is removed, and a copy of the feature map $F_5$ output by the main network is directly copied to the multi-scale feature fusion module.
\item w/o MSF: The multi-scale feature fusion module MSFFM is removed, and the transmitted features are directly spliced with the upsampling features in a way of addition (to ensure the consistency of channel numbers).
\item w/o MSR: The multi-scale residual convolution module MSRCM is replaced with the original residual structure.
\end{itemize}

\begin{table}[H]
\centering
\caption{Experimental results of DCFFSNet ablation}
\label{tab:ablation}
\begin{tabular}{llcc}
\toprule
Dataset & Net & IoU(\%) & Dice(\%) \\
\midrule
\multirow{4}{*}{ISIC2018} & w/o DS & 81.1 $\pm$ 0.4 & 87.3 $\pm$ 0.5 \\
& w/o MSF & 82.8 $\pm$ 0.1 & 88.9 $\pm$ 0.1 \\
& w/o MSR & 83.2 $\pm$ 0.2 & 89.5 $\pm$ 0.2 \\
& DCFFSNet (Ours) & \textbf{83.5 $\pm$ 0.3} & \textbf{90.0 $\pm$ 0.2} \\
\midrule
\multirow{4}{*}{DSB2018} & w/o DS & 83.7 $\pm$ 0.2 & 89.9 $\pm$ 0.2 \\
& w/o MSF & 84.8 $\pm$ 0.1 & 91.1 $\pm$ 0.1 \\
& w/o MSR & 85.0 $\pm$ 0.1 & 91.5 $\pm$ 0.1 \\
& DCFFSNet (Ours) & \textbf{85.4 $\pm$ 0.1} & \textbf{92.1 $\pm$ 0.1} \\
\midrule
\multirow{4}{*}{MoNuSeg} & w/o DS & 63.1 $\pm$ 1.1 & 77.2 $\pm$ 1.0 \\
& w/o MSF & 64.2 $\pm$ 0.7 & 78.3 $\pm$ 0.7 \\
& w/o MSR & 66.5 $\pm$ 0.6 & 79.2 $\pm$ 0.6 \\
& DCFFSNet (Ours) & \textbf{67.2 $\pm$ 0.9} & \textbf{79.7 $\pm$ 0.9} \\
\bottomrule
\end{tabular}
\end{table}

The DSCRIM, MSFFM, and MSR modules designed in this section have all positively impacted the network. For instance, in the ISIC dataset, removing the DSCRIM module resulted in a 2.4\% decrease in the Dice coefficient score and a 2.7\% decrease in the IoU score of DCFFSNet. Removing the MSFFM module led to a 0.7\% decrease in the Dice coefficient score and a 1.1\% decrease in the IoU score. Replacing the MSR module resulted in a 0.3\% decrease in the Dice coefficient score and a 0.5\% decrease in the IoU score.

\begin{figure}[H]
\centering
\includegraphics[width=0.8\textwidth]{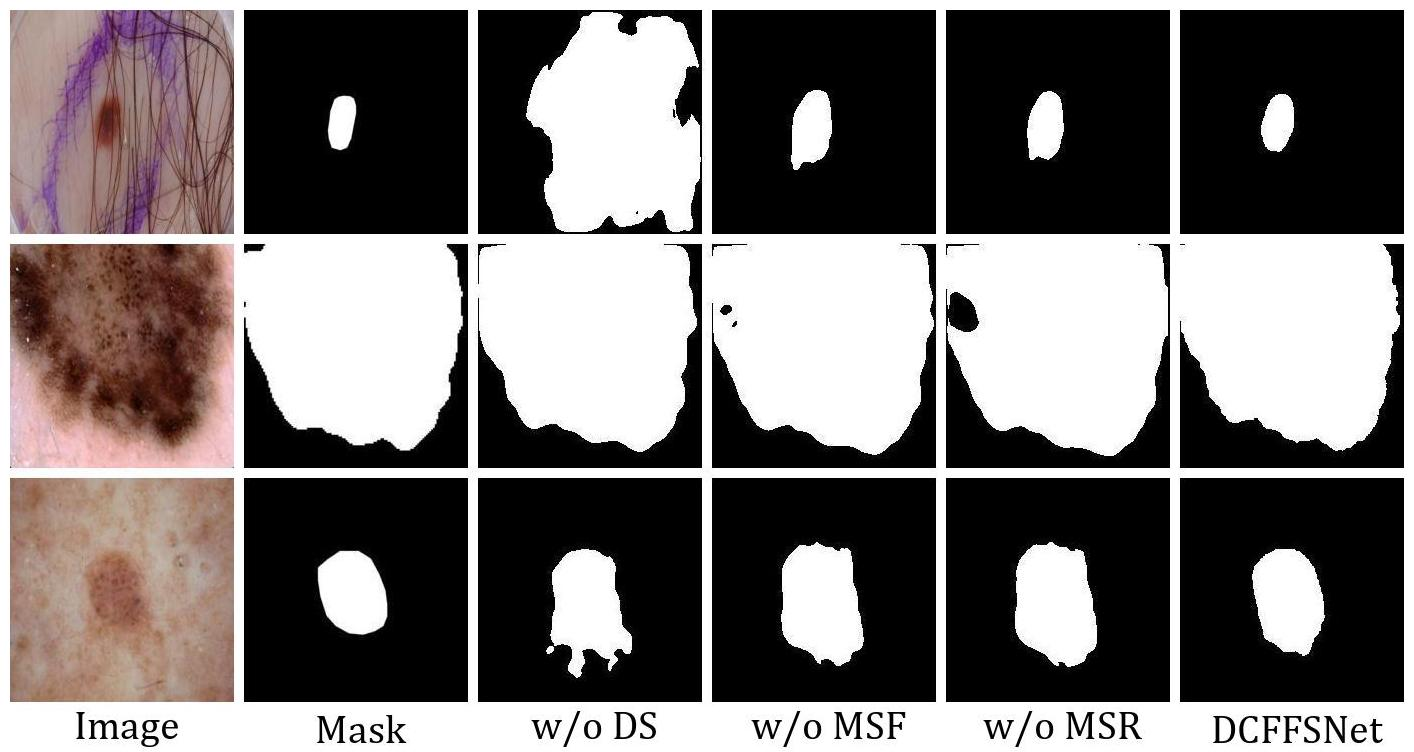}
\caption{Ablation experiment results of ISIC2018 dataset}
\label{fig:ablation_isic}
\end{figure}

\begin{figure}[H]
\centering
\includegraphics[width=0.8\textwidth]{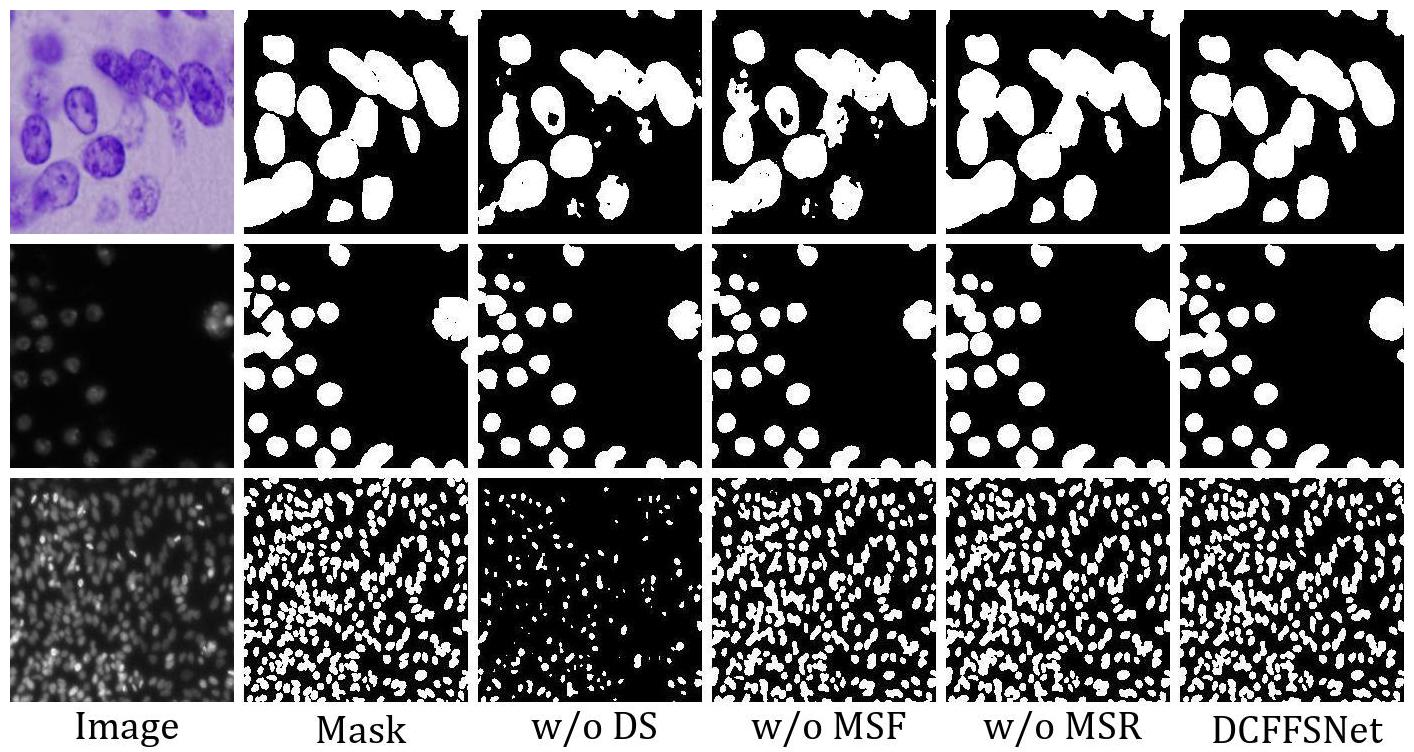}
\caption{Ablation experiment results of DSB2018 dataset}
\label{fig:ablation_dsb}
\end{figure}

\begin{figure}[H]
\centering
\includegraphics[width=0.8\textwidth]{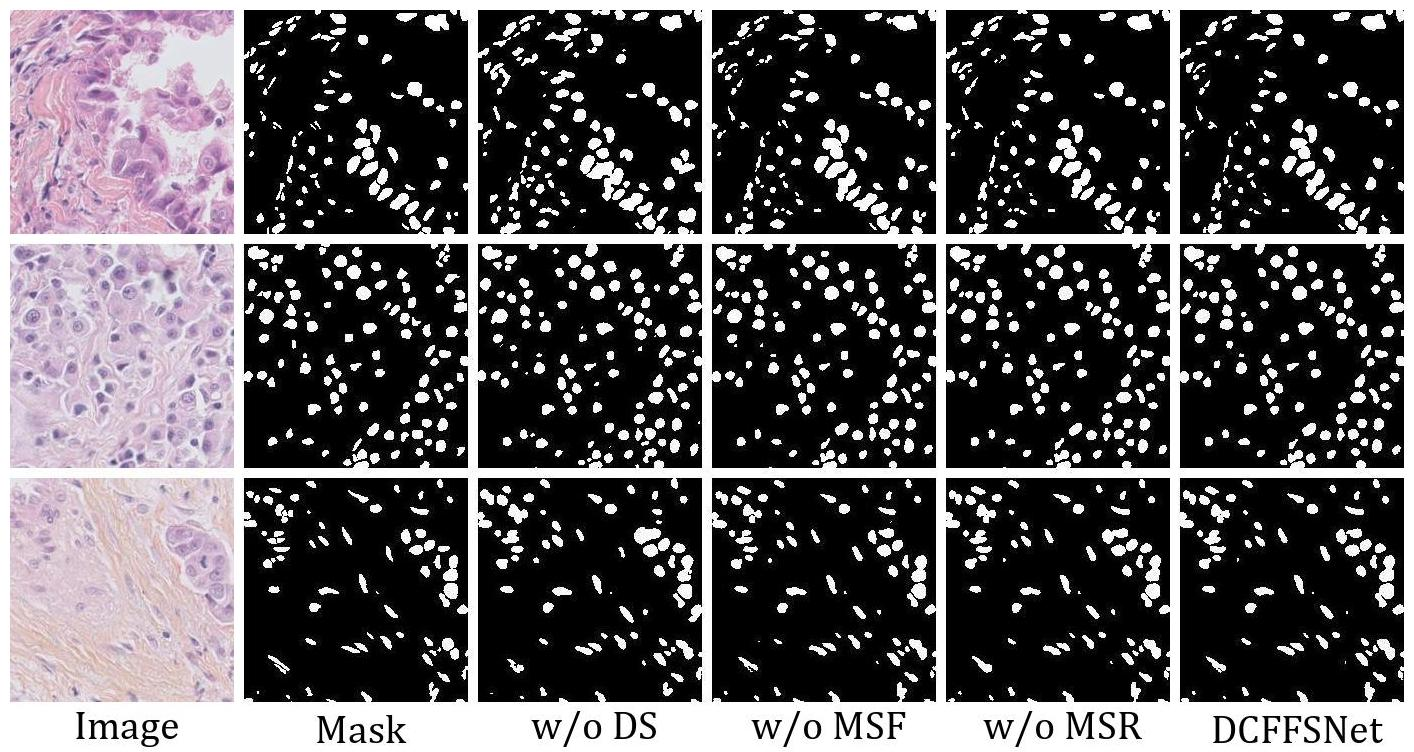}
\caption{Results of MoNuSeg dataset ablation experiment}
\label{fig:ablation_monuseg}
\end{figure}

Figures~\ref{fig:ablation_isic},~\ref{fig:ablation_dsb} and~\ref{fig:ablation_monuseg} show the visual results of ablation experiments.

The visualization results from the ablation experiments clearly show that the collaborative effects of various modules have different impacts on the model proposed in this chapter. The deep supervision connectivity representation injection module significantly influences the final model results, while DSCRIM has a substantial impact on both overall feature acquisition and connectivity acquisition. The multi-scale feature fusion module (MSFFM) has a significant effect on the internal topology. By integrating connectivity and feature scales, removing it would focus more on feature scale information, leading to the destruction of both internal and edge topologies. After the collaborative effects of multiple modules, the model proposed in this chapter not only ensures segmentation performance but also accurately determines boundary positions while maintaining internal structural consistency, ultimately achieving effective segmentation of medical images.

In addition to the ablation experiments mentioned above, corresponding experiments were also carried out on the weight allocation of the loss function used in this paper. Specifically, the weight of $L(output1)$ was adjusted, and the results are shown in Table~\ref{tab:weight_ablation}. In which $w = 0$ means that the weight is cancelled, that is, the deep supervision method is no longer used to inject connectivity scale features.

\begin{table}[H]
\centering
\caption{Weight ablation results of L(output1)}
\label{tab:weight_ablation}
\begin{tabular}{llcc}
\toprule
Dataset & Evaluating W & IoU(\%) & Dice(\%) \\
\midrule
\multirow{4}{*}{ISIC2018} & 0 & 81.7 $\pm$ 0.2 & 88.1 $\pm$ 0.2 \\
& 0.1 & 82.7 $\pm$ 0.3 & 88.8 $\pm$ 0.3 \\
& 0.2 & \textbf{83.5 $\pm$ 0.3} & \textbf{90.0 $\pm$ 0.2} \\
& 0.3 & 83.1 $\pm$ 0.2 & 89.5 $\pm$ 0.3 \\
\midrule
\multirow{4}{*}{DSB2018} & 0 & 84.9 $\pm$ 0.1 & 91.5 $\pm$ 0.1 \\
& 0.1 & 85.2 $\pm$ 0.1 & 91.9 $\pm$ 0.1 \\
& 0.2 & \textbf{85.4 $\pm$ 0.1} & \textbf{92.1 $\pm$ 0.1} \\
& 0.3 & 85.1 $\pm$ 0.1 & 91.7 $\pm$ 0.1 \\
\midrule
\multirow{4}{*}{MoNuSeg} & 0 & 64.8 $\pm$ 0.8 & 78.0 $\pm$ 0.8 \\
& 0.1 & 65.7 $\pm$ 0.9 & 78.5 $\pm$ 0.9 \\
& 0.2 & \textbf{67.2 $\pm$ 0.9} & \textbf{79.7 $\pm$ 0.9} \\
& 0.3 & 66.2 $\pm$ 0.9 & 78.7 $\pm$ 0.9 \\
\bottomrule
\end{tabular}
\end{table}

The results show that the model performs best when the weight of $L(output1)$ is 0.2. However, after the use of deep supervision to inject connectivity scale features is cancelled, the performance of the model decreases significantly, which also reflects the important role played by DSCRIM module in the model to some extent.

\section{Discussion}

The proposed algorithm excels in preserving internal topological structures and smoothing edge regions, effectively capturing target details and generating smooth segmentation boundaries; however, it demonstrates limited overall metric improvements (e.g., Dice coefficient/IoU) compared to state-of-the-art methods, likely due to insufficient global consistency focus while optimizing local details and potential slight accuracy reduction from connectivity constraints despite enhanced topological integrity. Future research should prioritize balanced local-global optimization through enhanced consistency mechanisms, develop parameterized strategies for dynamically adapting connectivity strength across scenarios, and integrate connectivity with advanced techniques like attention mechanisms or multi-scale feature fusion to comprehensively boost performance.

\section{Conclusion}

This paper proposes DCFFSNet (Deep Connectivity Feature Fusion Separation Network), a novel medical image segmentation framework that addresses the limitations of existing connectivity-based methods through systematic feature space decoupling. By introducing a three-layer architecture comprising the Deeply Supervised Connectivity Representation Injection Module (DSCRIM), Multi-Scale Feature Fusion Module (MSFFM), and Multi-Scale Residual Convolution Module (MSRCM), our approach effectively quantifies and balances the relative strength between connectivity features and multi-scale features. Experimental results on ISIC2018, DSB2018, and MoNuSeg datasets demonstrate superior performance compared to state-of-the-art methods, with improvements of up to 1.3\% in Dice coefficient and 1.2\% in IoU. The proposed method particularly excels in edge precision and regional consistency, offering significant advantages for clinical applications requiring accurate boundary delineation and internal structure preservation.

\section*{Acknowledgments}

We extend our sincere gratitude to: the International Skin Imaging Collaboration (ISIC) for the ISIC 2018 dataset (2,594 dermoscopic images, resolution: 542×718 to 4,499×6,748 pixels), whose precise lesion annotations served as a crucial benchmark for validating cross-scale feature fusion algorithms in our skin lesion segmentation model; Kaggle for the DSB2018 dataset (670 microscopy images, resolution: 256×256 to 1,024×1,024 pixels), whose densely distributed nuclei and complex backgrounds optimized tiny target detection capabilities in our nuclear instance segmentation framework; and the U.S. National Institutes of Health (NIH) for the MoNuSeg dataset (44 histopathology images at 1,000×1000 resolution), whose high-accuracy nuclear boundary annotations played a pivotal role in developing the boundary-aware loss function and enhancing contour segmentation precision---we hereby express our profound appreciation to all institutions advancing open science.

\section*{Declarations}

\begin{itemize}
  \item \textbf{Funding}: Not applicable.
  \item \textbf{Conflict of interest/Competing interests}: The authors declare no competing interests.
  \item \textbf{Ethics approval and consent to participate}: Not applicable.
  \item \textbf{Consent for publication}: Not applicable.
  \item \textbf{Availability of data and materials}: The ISIC2018 dataset is publicly available at \url{https://challenge.isic-archive.com/data/}. The DSB2018 dataset is available at \url{https://www.kaggle.com/c/data-science-bowl-2018}. The MoNuSeg dataset is available at \url{https://monuseg.grand-challenge.org/}.
  \item \textbf{Materials availability}: Not applicable.
  \item \textbf{Code availability}: The implementation code is available from the corresponding author on reasonable request.
  \item \textbf{Authors' contributions}: XY conceived the idea and designed the architecture. RT and MZ conducted the experiments. JQ supervised the project and provided critical revisions. All authors contributed to writing the manuscript.
\end{itemize}


\end{document}